\documentclass{article}
\usepackage[main,final]{neurips_2026}

\usepackage[T1]{fontenc}
\usepackage[utf8]{inputenc}
\usepackage{tgtermes}          % TeX Gyre Termes (pdfLaTeX-compatible)
\usepackage{microtype}

\usepackage{amsmath,amssymb,amsthm}
\usepackage{mathtools}

\usepackage{graphicx}
\usepackage{booktabs}
\usepackage{multirow}
\usepackage{multicol}
\usepackage{tabularx}
\usepackage{array}
\usepackage[table]{xcolor}

\usepackage[colorlinks=true,
            linkcolor=black,
            citecolor=blue!60!black,
            urlcolor=blue!60!black,
            hypertexnames=false,
            backref=page]{hyperref}
\usepackage{url}

\usepackage{algorithm}
\usepackage{algpseudocode}

\usepackage{listings}
\usepackage{xspace}
\usepackage{enumitem}
\usepackage{subcaption}
\usepackage{wrapfig}
\usepackage[capitalise]{cleveref}

\newcommand{\ours}{\textsc{AS$^2$}\xspace}
\newcommand{\perceptiononly}{\textsc{Perception-Only}\xspace}
\newcommand{\pipeline}{\textsc{Pipeline}\xspace}
\newcommand{\satnet}{\textsc{SATNet}\xspace}
\newcommand{\rrn}{\textsc{RRN}\xspace}
\newcommand{\neurasp}{\textsc{NeurASP}\xspace}
\newcommand{\deepproblog}{\textsc{DeepProbLog}\xspace}
\newcommand{\scallop}{\textsc{Scallop}\xspace}
\newcommand{\ltn}{\textsc{LTN}\xspace}
\newcommand{\anesi}{\textsc{A-NeSI}\xspace}
\newcommand{\slashmethod}{\textsc{SLASH}\xspace}
\newcommand{\pbcs}{\textsc{PBCS}\xspace}

\newcommand{\vsudoku}{Visual~Sudoku\xspace}
\newcommand{\mnistAdd}{MNIST\textsubscript{Add}\xspace}

\newcommand{\symacc}{\textsc{Sym-Acc}\xspace}
\newcommand{\boardacc}{\textsc{Board-Acc}\xspace}
\newcommand{\csr}{\textsc{CSR}\xspace}

\newcommand{\digitacc}{\textsc{Digit-Acc}\xspace}
\newcommand{\sumacc}{\textsc{Sum-Acc}\xspace}

\newcommand{\na}{N/A}

\newcommand{\best}[1]{\ifmmode\boldsymbol{#1}\else\textbf{#1}\fi}
\newcommand{\second}[1]{\underline{#1}}

\newcommand{\up}{\texorpdfstring{$\uparrow$}{(up)}}

\newcommand{\eg}{\emph{e.g.},\xspace}
\newcommand{\ie}{\emph{i.e.},\xspace}

\newcommand{\cleverHansThree}{CLEVR-Hans3\xspace}
\newcommand{\cleverHansSeven}{CLEVR-Hans7\xspace}
\newcommand{\sceneacc}{\textsc{Scene-Acc}\xspace}
\newcommand{\cnnDefault}{\textsc{CNN-Default}\xspace}
\newcommand{\cnnXIL}{\textsc{CNN-XIL}\xspace}
\newcommand{\nesyCL}{\textsc{NeSy-CL}\xspace}
\newcommand{\nesyXIL}{\textsc{NeSy-XIL}\xspace}

\hypersetup{allcolors=black}

\newcolumntype{C}[1]{>{\centering\arraybackslash}p{#1}}

\title{AS$^2$ - \underline{A}ttention-Based \underline{S}oft \underline{A}nswer \underline{S}ets: An End-to-End Differentiable Neuro-\emph{Soft}-Symbolic Reasoning Architecture}

\author{%
  Wael Abd-Almageed \\
  Holcombe Department of Electrical and Computer Engineering \\
  Clemson University \\
  \texttt{wabdalm@clemson.edu}
}

\begin{document}

\maketitle

\begin{abstract}
Neuro-symbolic systems typically couple a neural perception module with a
discrete symbolic solver, where exact constraint inference is intractable at
training time and a discrete solver must be invoked at inference time, precluding end-to-end optimization of the
model.
We introduce \ours{} (Attention-Based Soft Answer Sets), a fully
differentiable neuro-symbolic architecture that replaces the discrete solver
with a continuous approximation of the Answer Set Programming (ASP)
immediate-consequence operator $T_P$.
\ours{} maintains per-position probability distributions over a finite symbol
domain throughout the forward pass and trains end-to-end by minimizing the
fixed-point residual of a probabilistic lift of $T_P$, thereby
\emph{differentiating through the constraint check} without invoking an
external solver during either training or inference.
On spatial constraint-satisfaction tasks, \ours{} replaces
conventional positional embeddings entirely, and problem structure is encoded
through constraint-group membership embeddings derived directly from the
declarative ASP specification, making the model agnostic to arbitrary
position indexing.
On Visual Sudoku, \ours{} achieves 99.89\% raw cell accuracy and, after greedy decoding, 100\% constraint satisfaction, without invoking any external solver.
On CLEVR-Hans,  \ours{} achieves $99.94\%$ on
\cleverHansThree and $87.77\%$ on \cleverHansSeven.
These results demonstrate that a soft differentiable fixpoint operator,
combined with constraint-aware attention and declarative constraint
specification, can match or exceed pipeline and solver-based neuro-symbolic
systems while maintaining full end-to-end differentiability.
\end{abstract}

% ============================================================
% Section 1: Introduction
% ============================================================
\section{Introduction}
\label{sec:introduction}

The integration of neural perception and symbolic reasoning has
received considerable attention in recent years~\citep{garcez2019neural, manhaeve2018deepproblog, li2023scallop}. However, neuro-symbolic pipelines either rely on intractable exact inference at training time or invoke a discrete symbolic solver at inference time, with the boundary between perception and reasoning components imposing a structural cost on both regimes. Perceptual constraint-satisfaction problems (P-CSPs)~\citep{wang2019satnet, yang2020neurasp, mulamba2024perception} require 
mapping raw sensor data (\eg images) to discrete
symbolic assignments that collectively satisfy a set of declarative
constraints (\eg physical laws, etc.).
Neural models excel at extracting high-level representations from raw,
high-dimensional inputs, while symbolic reasoners excel at enforcing
global consistency through declarative constraints.
The challenge of combining these two capabilities arises from the
incompatibility of their computational registers, since neural inference
operates over continuous distributions while symbolic reasoning requires
discrete variable assignments that satisfy hard logical constraints.

The dominant paradigm in neuro-symbolic artificial intelligence (AI) addresses P-CSPs through a
\emph{pipeline} architecture that decouples perception from reasoning, in which a neural network first processes inputs, then a
discrete symbolic solver is invoked to find a globally consistent assignment.
Answer Set Programming (ASP)~\citep{brewka2011answer, gelfond1988stable} is the predominant
symbolic formalism employed in such pipelines. ASP is a mature declarative logic-programming paradigm
whose stable-model semantics support default negation, choice rules, and optimization statements,
enabling compact specification of complex combinatorial constraints.
Clingo~\citep{gebser2014clingo} is the most widely used ASP solver, while
Constraint Programming SAT (CP-SAT) solvers~\citep{mulamba2024perception} provide a propositional alternative when the
constraint vocabulary does not require full ASP expressiveness.

Pipeline architectures introduce two structural costs that motivate this work. First, although several pipeline systems are end-to-end differentiable at training time through probabilistic stable models~\citep{yang2020neurasp}, weighted model counting~\citep{manhaeve2018deepproblog}, or differentiable provenance semirings~\citep{li2023scallop}, the underlying exact inference is in general computationally intractable (\#P-hard for weighted model counting), forcing approximations that compromise the quality of the constraint signal at scale. Second, all of these systems still invoke a discrete solver at inference time, which precludes deployment in latency-sensitive or resource-constrained settings and prevents end-to-end optimization of the inference path itself. \pbcs~\citep{mulamba2024perception} additionally relies on a post-hoc no-good surrogate rather than a direct gradient through the constraint check. A confident perception error
(\eg misclassification) can therefore render the symbolic grounding unsatisfiable, and the rich uncertainty encoded in the learned representations of the perception module is not exposed to the reasoning component during inference.

Alternative approaches attempt end-to-end differentiability by replacing
the discrete solver with differentiable relaxation.
\satnet~\citep{wang2019satnet} embeds a semi-definite programming (SDP)-relaxed Maximum Satisfiability (MaxSAT) layer as a
neural module, and \rrn~\citep{palm2018recurrent} learns iterative
message passing without any explicit symbolic component.
However, these systems either lack formal constraint guarantees or rely on relaxations that are not sound with respect to the constraint semantics, and they do not generalize to the expressive ASP constraint vocabulary.

We introduce \ours{} (\underline{A}ttention-Based \underline{S}oft \underline{A}nswer \underline{S}ets), a fully differentiable neuro-\emph{soft}-symbolic architecture built on three coupled design principles. (i) Rather than committing to discrete symbol assignments before reasoning, the model maintains continuous probability distributions over a finite symbolic vocabulary throughout the entire forward pass and enforces constraints via a differentiable loss derived from the fixed-point residual of a probabilistic lift of the ASP immediate-consequence operator $T_P$~\citep{van_emden1976semantics,takemura2024differentiable}, which replaces set-theoretic intersection with element-wise products and measures violation as the squared distance between the current distribution and its image under $T_P$, yielding a loss that is zero iff the prediction is a one-hot assignment satisfying all constraints and that admits no degenerate minimum at the uniform distribution, and gradients flow from constraint satisfaction directly into the perception encoder, and no external solver is invoked at training or inference time. (ii) On spatial constraint-satisfaction tasks, \ours{} dispenses with conventional positional embeddings entirely, where in contrast to neuro-symbolic Transformers that rely on learned or sinusoidal position encodings~\citep{mcleish2024transformers}, problem structure is encoded through \emph{constraint-group membership embeddings} read off the declarative ASP specification, making token representations invariant to permutations of the position index and grounded in the logical structure of the problem. (iii) The probabilistic $T_P$ surrogate is a structural translation of the symbolic operator into continuous arithmetic rather than an additional free-parameter layer, so its topology is fixed by the logic program while operator parameters are either fixed (explicit variant) or initialized from the rules and refined end-to-end (learned variant), so full differentiability does not collapse \ours{} into an unconstrained learned model, and the reasoning pathway is dictated by constraint semantics rather than learned from examples. This paper makes the following contributions.
\begin{itemize}[leftmargin=*, noitemsep, topsep=0pt]
  \item A fully differentiable neuro-\emph{soft}-symbolic architecture that replaces discrete ASP solvers with a soft probabilistic $T_P$ operator,
    enabling end-to-end training with constraint feedback.
  \item Constraint-group membership embeddings as a
    positional-embedding-free mechanism for encoding problem structure in
    attention-based models.
\end{itemize}
\nopagebreak[4]\vspace{0pt}

% ============================================================
% Section 2: Related Work
% ============================================================
\section{Related Work}
\label{sec:related_work}

We organize related work into three major constraint-solving families of methods that involve neural perception. The first family, pipeline neuro-symbolic systems, couples a neural perception module to a discrete symbolic solver and typically invokes the solver at inference time. The second family, end-to-end differentiable neuro-symbolic systems, replaces the discrete solver with a differentiable relaxation. The third family, perception-based constraint solving methods, supervises a neural perception module jointly with a classical constraint solver using calibration feedback computed during training, and invokes the solver again at inference.

\paragraph{Pipeline Neuro-Symbolic Systems.}\label{sec:rw_pipeline}
By far, the dominant paradigm in neuro-symbolic artificial intelligence (AI) couples a neural perception
module with a classical symbolic solver. Several systems in this family are end-to-end differentiable at training time but still rely on a discrete solver or an intractable exact-inference procedure to compute the constraint-satisfaction signal.
\neurasp~\citep{yang2020neurasp} wraps neural network outputs as
probabilistic facts in an Answer Set Programming (ASP) program and computes gradients through probabilistic stable models. Training-time gradients therefore reach the perception encoder, but inference still invokes Clingo, and the exact stable-model count grows combinatorially with the program size.
\deepproblog~\citep{manhaeve2018deepproblog,manhaeve2021neural} attaches
neural predicates to a ProbLog engine and computes exact gradients
through weighted model counting. Weighted model counting is \#P-hard in general, so the framework does not scale beyond small programs without approximation, and stable-model semantics are not supported.
\scallop~\citep{li2023scallop} introduces differentiable Datalog with
provenance semirings. Its constraint vocabulary is limited to
stratified Datalog, which supports stratified default negation but excludes unstratified default negation, choice rules, and
optimization statements. At scale, the provenance circuits must be approximated by top-$k$ proofs, sacrificing soundness and completeness of the gradient signal.
\ltn~\citep{badreddine2022logic} grounds first-order logic formulae as
differentiable real-valued functions, yet the fuzzy-logic relaxation
discards crisp stable-model semantics, making formal verification
intractable.
The shared limitation of pipeline systems is therefore not a lack of training-time differentiability, but rather (i) reliance on a discrete solver at inference time, (ii) intractable or approximated exact inference for the training signal, and (iii) restricted constraint vocabularies that exclude full ASP.

\paragraph{End-to-End Differentiable Neuro-Symbolic Systems.}\label{sec:rw_e2e}
\satnet~\citep{wang2019satnet} embeds a differentiable semi-definite programming (SDP)-relaxed
MaxSAT layer as a neural module, enabling end-to-end training.
However, the SDP relaxation may certify
infeasible instances as satisfiable, and its $O(n^3)$ complexity
prevents scaling to large constraint systems.
\rrn~\citep{palm2018recurrent} learns iterative message passing on a
fully-connected cell graph, solving Sudoku without any explicit symbolic
component. It serves as a strong neural baseline, yet offers no formal
constraint guarantees and does not generalize to new constraint families.
\anesi~\citep{van2023anesi} addresses the intractability of exact
inference in \deepproblog by introducing approximate neuro-symbolic
inference, at some cost to the tightness of gradient estimates.
\slashmethod~\citep{skryagin2022slash} combines probabilistic circuits
with neural ASP for tractable inference via sum-product networks, yet
the coupling still relies on a classical solver at test time.

The work most closely related to ours is ~\citet{takemura2024differentiable}, who propose differentiable
constraint learning via the immediate-consequence operator for neural
ASP.
Their formulation provides the probabilistic lift of $T_P$ that we adopt as
the foundation for our constraint loss.
However, \citet{takemura2024differentiable} do not address the
architectural question of how to propagate constraint information across
positions without positional embeddings, and do not incorporate
inference-time constraint propagation.
\ours{} builds on their $T_P$ formulation and extends it with
constraint-group membership embeddings, a multi-layer Transformer
reasoning module, and greedy constrained decoding, yielding a complete
end-to-end system for perceptual constraint-satisfaction problems.

\paragraph{Perception-Based Constraint Solving.}\label{sec:rw_pbcs}
\citet{mulamba2024perception} introduce a predict-then-optimize
framework (\pbcs) that trains a Cell-CNN digit
classifier with calibration feedback derived from a CP-SAT solver,
which is also invoked at inference time.
On MNIST-based Visual Sudoku, \pbcs achieves 99.4\% grid accuracy, by
exploiting the fact that the distribution over digit labels, when
properly calibrated, provides enough soft evidence for CP-SAT to
back-propagate constraint violations into the perception loss via a
no-good-aware surrogate.
However, \pbcs has three structural limitations of the pipeline
paradigm.
First, the no-good surrogate is a post-hoc approximation of the true constraint
signal rather than a direct gradient through the constraint check.
Second, the constraint vocabulary is limited to \textsc{AllDifferent}. Extending to recursive rules, default negation, or optimization
statements would require a new solver entirely.
Third, \pbcs applies only when a complete, sound CP model exists for the
target task, and a CP-SAT solver is invoked at inference time.
\ours{} addresses constraint feedback through  differentiable layers. Gradients flow end-to-end through the
differentiable $T_P$ residual loss, and inference is solver-free. Extending the formal coverage to default negation, choice rules, and optimization statements is left to future work.

\paragraph{Positional Encodings in Neuro-Symbolic Transformers.}\label{sec:rw_positional}
\citet{mcleish2024transformers} encode Row-Column-Box membership as fixed binary features prepended to each token, and constraint-graph relative positional encodings~\citep{palm2018recurrent} bias self-attention by the graph distance between constrained pairs.
\citet{yang2023learning} incorporate constraint-graph structure into a
Recurrent Transformer for CSPs, demonstrating competitive performance on Visual Sudoku
using learned positional embeddings, and \citet{xu2025selfsupervised}
extend this line of work with a self-supervised refinement objective that
generalizes to out-of-distribution problem sizes.
\citet{kang2025benchmark} introduce a pixel-sequence benchmark explicitly
designed to evaluate structural reasoning that cannot be resolved by sequential
positional cues alone.
Unlike these approaches, \ours{} derives its token representations directly from the declarative ASP specification, so the embedding topology is a function of the logic program rather than a hand-crafted geometric prior, and the same compilation procedure applies unchanged to any P-CSP that can be expressed in ASP. When the constraint graph is dynamic or instance-dependent (\eg in scheduling
or planning domains where which variables share a constraint varies per instance), the constraint-group membership embeddings can be recompiled from the instance-specific ground program, since the compilation maps each ground atom to the set of rules in which it participates.

\ours{} takes a fundamentally different approach,
replacing conventional positional embeddings entirely with
\emph{constraint-group membership embeddings}, where each token's
representation is enriched by the sum of learned embeddings indexed by
the constraint groups to which the token belongs (\eg row, column, and box
for Sudoku).
This design directly encodes the logical structure of the problem into
the token representations and makes the model invariant to any
re-indexing of positions that preserves constraint-group membership.
To our knowledge, \ours{} is the first neuro-symbolic Transformer that
achieves competitive results on Visual Sudoku without any form of
conventional positional encoding.

% ============================================================
% Section 3: Theoretical Framework
% ============================================================

\begin{figure}[h]
  \centering
  \includegraphics[width=\linewidth]{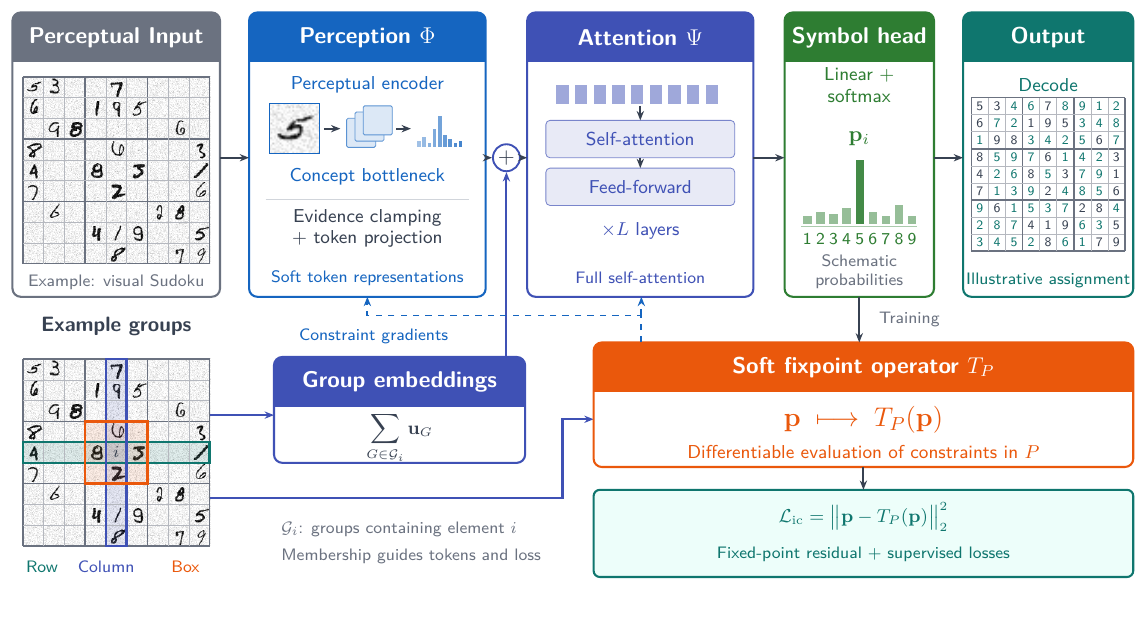}
  \caption{%
    \textbf{\ours{}} processes inputs through a neural perception module and transformer encoder, producing distributions that feed a differentiable soft fixpoint operator that reasons over symbolic concept rules.
  }
  \label{fig:architecture}
  \vspace{-.15in}
\end{figure}

\section{Attention-Based Soft Answer Sets (\ours{})}
\label{sec:method}

% ─────────────────────────────────────────────────────────────
\subsection{Problem Formulation and Design Principles}
\label{sec:problem}
% ─────────────────────────────────────────────────────────────

In \emph{perceptual constraint-satisfaction} problems (P-CSPs), a model
observes $n$ raw perceptual signals $\mathbf{X} = (X_1, \dots, X_n)$ and
must assign each sample $i$ a symbol from a finite domain
$\Sigma = \{s_1, \dots, s_k\}$, subject to a set of declarative integrity
constraints $\mathcal{C}$ specified as a logic program $P$ over ground
atoms drawn from a Herbrand structure over $\Sigma$. A binary evidence
mask $\mathbf{m} \in \{0,1\}^n$ identifies samples whose assignments
are directly observable ($m_i = 1$, \emph{evidence}) versus samples
that must be inferred ($m_i = 0$, \emph{latent}). The ground-truth
assignment is $\mathbf{y} \in \Sigma^n$ and the task is to predict
$\hat{\mathbf{y}} \in \Sigma^n$ that (i)~agrees with all evidence
samples and (ii)~is an answer set of $P$, satisfying every constraint
in $\mathcal{C}$. This formulation is generic, since any task in which
perceptual evidence must be jointly mapped to a finite symbolic
configuration that obeys declarative constraints fits within it.

%\blue{ As illustrated in \cref{fig:architecture}, \ours{} addresses limitations of classic neuro-symbolic architectures that invoke discrete solvers at inference time by using soft representations end-to-end. Instead of committing to discrete symbols, \ours{} maintains per-sample probability distributions $\mathbf{p} \in [0,1]^{n \times k}$ over $\Sigma$ and differentiates through the constraint check using a probabilistic lift of the immediate-consequence operator. The solver is replaced by a differentiable loss that drives $\mathbf{p}$ to be simultaneously consistent with the perceptual evidence \emph{and} with the constraints in $\mathcal{C}$, allowing gradients to flow from reasoning back into perception. At inference, the predicted assignment is recovered by $\hat{y}_i = \operatorname{argmax}_s p_{i,s}$, without external solvers.}

As illustrated in \cref{fig:architecture}, \ours{} is a neuro-symbolic constraint-satisfaction architecture
in which perception, attention, and
soft-symbolic reasoning layers are coupled through a single differentiable computation graph. \ours{} realizes three fundamental properties. First, gradients must flow from symbolic reasoning to perception, so perception is shaped by reasoning rather than learned in isolation. Second, integrity constraints are declaratively specified. Third, logic operators are represented as differentiable layers, and therefore no external solver is needed at either training or inference time.

% ─────────────────────────────────────────────────────────────
\subsection{\ours{} Architecture}
\label{sec:architecture}
% ─────────────────────────────────────────────────────────────

%\revadd{\ours{} comprises three task-specific $T_P$-style layers, namely mutual exclusivity with cardinality constraints for Sudoku, existential conjunction for CLEVR-Hans, and arithmetic sum constraints for MNIST Addition.} \revadd{These are separately hand-derived constructions, not a general ASP compiler, and do not implement general default negation, choice rules, disjunction, optimization, or recursive stable-model semantics.}

\paragraph{Declarative Constraint Specification.}
\label{sec:declarative}
Constraints are \emph{declaratively} specified in standard
ASP syntax~\citep{gebser2014clingo}, using
notation compatible with symbolic solvers (\eg Clingo.) The user specifies what
constitutes a valid assignment, and the framework  compiles
each integrity constraint $\mathrel{:\!-}\,\phi$ into the soft
computation graph. Compilation
proceeds by extracting from $\mathcal{C}$ a collection of
\emph{constraint groups} $\mathcal{G} = \{G_1, \dots, G_M\}$, where each
$G \subseteq \{1,\dots,n\}$ is a subset of samples whose assignments
must be jointly consistent with respect to a particular symbol-level condition. The collection $\mathcal{G}$ is the bridge between the
symbolic specification and the neural representation.

\paragraph{Perception.}
Perception maps raw perceptual signal $X_i$ to a
continuous embedding $\mathbf{h}_i \in \mathbb{R}^{d_\text{model}}$ and then to pre-reasoning logits
$\mathbf{z}_i^\text{pre} \in \mathbb{R}^k$ through a \emph{concept bottleneck head}~\citep{koh2020concept}, which is a single linear layer that
provides an interpretable interface between perception and reasoning.

\paragraph{Evidence Clamping.}
Evidence clamping replaces the soft
distribution of observed samples ($m_i = 1$) with the one-hot ground-truth label during training, or with the perception module's $\operatorname{argmax}$ prediction during inference, anchoring the latent representations at known samples. The generated $k$-dimensional soft distributions are then projected back to
 $d_\text{model}$ to form the input tokens of the reasoning module.

\paragraph{Constraint-Group Membership.}
Instead of using conventional positional embeddings to encode spatial structure, \ours{} encodes problem
structure through \emph{constraint-group membership embeddings}. Each token's representation is enriched by the element-wise sum of learned embedding vectors indexed by the constraint groups in $\mathcal{G}$ to
which the token belongs. This design  encodes the logical
structure of the constraint program into the token representations, so
that two samples share a high degree of representational similarity if
and only if they participate in the same constraint groups. The model is
therefore invariant to any re-indexing of samples that preserves
constraint-group membership, and the embeddings do not impose any
arbitrary sequential ordering on the problem variables.

\paragraph{Reasoning.}
The $n$ tokens, enriched with constraint-group membership embeddings,
are processed by a multi-layer Transformer
encoder~\citep{vaswani2017attention} with pre-norm (LayerNorm before
attention and feedforward), Gaussian Error Linear Unit (GELU) activation, and full self-attention.
Self-attention is the natural operator at this stage because every
sample must reconcile its own perceptual evidence with the assignments
at every other sample that shares a constraint group, and a single
attention layer aggregates evidence across all samples simultaneously.
Stacking attention layers therefore allows the latent representations to
collectively converge toward a globally consistent assignment without
ever committing to a discrete symbol. The output head projects each
token to post-reasoning \emph{logits}
$\mathbf{z}_i^\text{post} \in \mathbb{R}^k$, which are passed through
\emph{softmax} to produce the per-sample distributions
$\mathbf{p} \in [0,1]^{n \times k}$.

\paragraph{Soft-Logic Layer.}
At this point we have soft per-sample
distributions that are coherent with the perceptual evidence but not yet
guaranteed to satisfy the integrity constraints in $\mathcal{C}$. We need a mechanism that pushes $\mathbf{p}$ toward configurations
that are simultaneously valid one-hot assignments \emph{and} consistent
with $\mathcal{C}$, while remaining differentiable so that gradients can
flow back through the reasoning and perception modules. The soft
fixpoint layer realizes this mechanism by lifting the
\emph{immediate-consequence operator} $T_P$ of the constraint program to probability
distributions. For a definite logic program $P$, the least fixed point of the classical $T_P$ is its unique answer set. A probability tensor $\mathbf{p}$
satisfies all constraints if and only if it is a fixed point of the
soft lift of $T_P$. Training therefore minimizes the residual
$\|\mathbf{p} - T_P(\mathbf{p})\|$ together with a supervised perception
loss, and constraint satisfaction emerges as a by-product of driving
$\mathbf{p}$ toward fixed points of $T_P$.
The same lifting recipe extends to qualitatively different rule families, as demonstrated in Section~\ref{sec:exp_clevr} for existential conjunctive rules.
% The remainder of this section defines $T_P$ formally, gives two concrete instantiations covering qualitatively different constraint families, and specifies the fixed-point loss used during training.

% ─────────────────────────────────────────────────────────────

\paragraph{Probabilistic Lift of the Immediate-Consequence Operator.} Let $P$ be a definite logic program and $T_P$ its
\emph{immediate-consequence operator}~\citep{van_emden1976semantics},
the function that maps a Herbrand interpretation $I$ to the set of atoms
whose bodies are satisfied under $I$. The least fixed point
$\text{lfp}(T_P)$ equals the unique minimal Herbrand model of $P$, and a
ground atom $a$ is a logical consequence of $P$ if and only if
$a \in \text{lfp}(T_P)$. The integrity constraints in $\mathcal{C}$ take
the form $\mathrel{:\!-}\,\phi$, which are violated whenever $\phi$
holds under the predicted assignment, and each such constraint induces
the constraint group $G \subseteq \{1,\dots,n\}$. Given per-element distributions $\mathbf{p} \in [0,1]^{n \times k}$ with $\sum_s p_{i,s}=1$ for every cell $i$, the probabilistic $T_P$ operator
applied to a mutual-exclusivity group $G \in \mathcal{G}$ is defined
element- and symbol-wise as shown in \cref{eq:tp},
\begin{equation}
  T_P(\mathbf{p})_{i,s}
  \;=\;
  p_{i,s} \cdot \prod_{j \in G,\, j \neq i}
  \bigl(1 - p_{j,s}\bigr),
  \qquad i \in G,\; s \in \Sigma.
  \label{eq:tp}
\end{equation}
Intuitively, element $i$ can \emph{claim} symbol $s$ only if it already
assigns probability mass to $s$ \emph{and} no competing element of $G$
does so. For normalized inputs and $|G|=k$, the fixed-point condition $\mathbf{p} = T_P(\mathbf{p})$ holds
if and only if $\mathbf{p}$ is a valid one-hot assignment in which every
group constraint is satisfied. The softmax outputs enforce this normalization, excluding the all-zero tensor from the admissible domain.

% ─────────────────────────────────────────────────────────────
\subsection{The $T_P$ Fixed-Point Loss}
\label{sec:tp_loss}
% ─────────────────────────────────────────────────────────────

Naive constraint penalties for exclusivity, such as the squared-sum loss
$(\sum_i p_{i,s} - 1)^2$~\citep{wang2019satnet}, have a degenerate
global minimum at the uniform distribution $p_{i,s} = 1/k$, which
satisfies no actual constraint. The \ours{} constraint loss avoids this
pathology by directly minimizing the fixed-point residual of $T_P$
across all constraint groups, as defined in \cref{eq:tp_loss},
\begin{equation}
  \mathcal{L}_{\text{ic}}(\mathbf{p})
  \;=\;
  \sum_{G \in \mathcal{G}}
  \sum_{i \in G} \sum_{s \in \Sigma}
  \bigl(p_{i,s} - T_P(\mathbf{p})_{i,s}\bigr)^2.
  \label{eq:tp_loss}
\end{equation}
At the uniform distribution,
$T_P(\mathbf{p})_{i,s} = (1/k)\cdot((k{-}1)/k)^{|G|-1} \neq 1/k$ for any
$k \geq 2$ and $|G| \geq 2$, so $\mathcal{L}_{\text{ic}} > 0$, where the
group size $|G|$ and domain size $k$ need not coincide in general.
Therefore, the degenerate minimizer of the squared-sum loss is
\emph{not} a minimizer of $\mathcal{L}_{\text{ic}}$. Conversely, at any
valid one-hot assignment the product in Equation~\eqref{eq:tp} collapses
to $p_{i,s} \cdot 1^{|G|-1} = p_{i,s}$, giving
$\mathcal{L}_{\text{ic}} = 0$. The entire computation is differentiable
and fully vectorized in log-space for numerical
stability~\citep{takemura2024differentiable}, so no solver is invoked
during either training or inference. An alternative arithmetic
instantiation of $T_P$ for global-sum constraints is described in
Appendix~\ref{sec:mnist_supp}.

% ============================================================
% Section 4: Experimental Evaluation
% ============================================================
\section{Experimental Evaluation}
\label{sec:experiments}

We apply \ours{} to three neuro-symbolic benchmarks spanning
different constraint types and input modalities. In this section, we discuss the results of applying \ours{} to Visual Sudoku~\citep{wang2019satnet} and CLEVR-Hans~\citep{stammer2021right}.
Visual Sudoku is governed by mutual-exclusivity rules over a fixed grid of grayscale digits.
For Sudoku, $T_P$ is fixed throughout training. CLEVR-Hans is governed by existential conjunctive rules over a variable-size object set in color 3-D scenes.
Its Explicit variant fixes the ground-truth rules, while its Learned-operator variant, used only for CLEVR-Hans, updates rule-initialized weights end-to-end.
 Across these benchmarks, \ours{} uses clue and blank-cell digit labels for Sudoku, scene-class and per-object attribute labels for CLEVR-Hans, and digit and sum labels for MNIST Addition.
Table~\ref{tab:rebuttal_supervision} in the supplementary material details this supplied information, which qualifies comparisons with published methods.
Full dataset statistics and \ours{} instantiations as well as results on MNIST Addition~\citep{li2023scallop} are deferred to { Appendices~\ref{sec:benchmark_stats}--\ref{sec:mnist_supp}}.

\subsection{Visual Sudoku}
\label{sec:exp_sudoku}
% ─────────────────────────────────────────────────────────────

A $9{\times}9$ Sudoku puzzle is rendered as a $252{\times}252$ grayscale
image by placing MNIST images in \emph{clue} cells and leaving blank  non-clue cells~\citep{wang2019satnet, yang2020neurasp}.
A model receives the full board image and must predict all 81 cell values
while satisfying three families of Sudoku constraints, requiring that
each row, each column, and each $3{\times}3$ box contains the digits
1--9 exactly once.
We use 9{,}000 / 1{,}000 / 1{,}000 boards for train / validation / test
with approximately 45 clue cells per board.
This benchmark requires genuine constraint
reasoning over 27 interacting constraint groups and cannot be solved by
perception alone. The instantiation of \ours{} for Visual Sudoku is described in detail in Appendix~\ref{sec:sudoku_supp}, as well as the complete declarative constraint specification (\cref{fig:asp_sudoku}).

\paragraph{Metrics.}
For Visual Sudoku, we report symbolic accuracy (\symacc), defined as the fraction of individual cells predicted correctly, board accuracy (\boardacc), defined as the fraction of boards in which \emph{all} 81 cells are correct, and constraint satisfaction rate (\csr), defined as the fraction of boards satisfying all row, column, and box constraints as measured algebraically from the model's output distributions.

\paragraph{Baselines.}
We compare against neuro-symbolic systems that  reported results on Visual Sudoku. \satnet~\citep{wang2019satnet} introduced the benchmark and embeds a differentiable SDP-relaxed MaxSAT layer. \pbcs~\citep{mulamba2024perception} applies perception-based CP-SAT constraint solving. We also include \pipeline~\citep{yang2020neurasp}, our reproduction of NeurASP-style perception followed by Clingo post-processing as a reference for solver-augmented inference. \rrn~\citep{palm2018recurrent} is evaluated on symbolic (digit-representation) Sudoku only and does not report Visual Sudoku results.

\paragraph{Results.}
As shown in Table~\ref{tab:results_vsudoku}, \ours{} achieves a cell accuracy (\symacc) of 99.89\%, reflecting
high accuracy on clue cells, where the prediction is anchored to the convolutional neural network (CNN) digit classifier's argmax, and
near-perfect accuracy on blank cells where the Transformer must infer the
correct digit from constraint propagation alone. The unaided board accuracy (raw argmax with no post-processing) is
95.60\%, reflecting residual constraint-satisfaction failures concentrated
in blank cells where the Transformer has not fully converged to a valid
fixed point.
Iterative $T_P$ refinement with $K{=}10$ steps
(Eq.\,\ref{eq:tp_refine}) raises the constraint satisfaction rate to
98.7\%, demonstrating that the trained distributions already encode
sufficient constraint information to recover approximately  {70.45\%} of the
argmax failures through soft propagation alone.
Greedy constrained decoding (Appendix~\ref{sec:sudoku_supp})
raises board accuracy to 100\% for the model-plus-decoder combination, improving on raw argmax by 4.4 percentage points, while achieving
a \csr of 100.0\% across all 1{,}000 test boards. Neither procedure calls
an external solver. The greedy decoder nevertheless makes discrete constraint-based assignments, and it does not include any form of backtracking or search, compared to traditional solver-augmented pipelines. \pbcs~\citep{mulamba2024perception} reports the highest
 board accuracy of 99.4\%  by coupling a
perception module with a full CP-SAT solver at inference time, although
it does not report constraint satisfaction metrics and requires a
complete constraint model at test time.
\rrn~\citep{palm2018recurrent} is evaluated on symbolic Sudoku only and is therefore not directly comparable on the perceptual task.
\ours{} achieves 100\% board accuracy and 100\% constraint satisfaction
without any external solver, surpassing all previously reported results.

% ────────────────────────────────────────────────────
% Table: Visual Sudoku
% ────────────────────────────────────────────────────
\vspace{-0.1in}
\begin{table}[h]
\centering
\caption{Results on \vsudoku. \best{Bold} = best, \second{underline} = second-best.
  $^\S$\pipeline \csr reflects Clingo post-processing, not learned satisfaction.
  We report three \ours{} inference regimes on the same 1{,}000 test boards: raw argmax of the Transformer output (pre-decoding), iterative $T_P$ refinement with $K{=}10$ steps applied to the soft distributions (Eq.\,\ref{eq:tp_refine}), and greedy constrained decoding (Appendix~\ref{sec:sudoku_supp}). None of the three regimes invokes an external solver. $^a$Evaluated on SATNet-V. $^b$Evaluated on RRN-V. Both Recurrent Transformer results use the ungrounded setting without clue-digit labels.}
\label{tab:results_vsudoku}
\setlength{\tabcolsep}{3.5pt}
\begin{tabular}{lcccc}
\toprule
\textbf{Method}
  & \textbf{Clues} & \symacc\up & \boardacc\up & \csr\up \\
\midrule
\satnet~\citep{wang2019satnet}
  & \na & \na & 63.2 & \na \\
\pipeline$^\S$~\citep{yang2020neurasp}
  & \na & 60.51 & 0.00 & 100.00 \\
\pbcs~\citep{mulamba2024perception}
  & \na & \na & \second{99.4} & \na \\
Recurrent Transformer$^a$~\citep{yang2023learning}
  & 31--42 & \na & 93.5 & \na \\
Recurrent Transformer$^b$~\citep{yang2023learning}
  & 17--34 & \na & 75.6 & \na \\
\midrule
\ours{} (raw argmax)
  & $\approx45$ & 99.89 & 95.60 & 95.60 \\
\ours{} ($+\,K{=}10$ $T_P$ refinement)
  & $\approx45$ & 99.89 & 98.70 & 98.70 \\
\ours{} ($+\,$greedy constrained decoding)
  & $\approx45$ & \best{100.0} & \best{100.0} & \best{100.0} \\
\bottomrule
\end{tabular}
\end{table}

% ─────────────────────────────────────────────────────────────
\subsection{CLEVR-Hans}
\label{sec:exp_clevr}
% ─────────────────────────────────────────────────────────────

CLEVR-Hans~\citep{stammer2021right} is a
visual concept-classification benchmark in which a model receives an RGB image of a synthetic 3-D scene containing between 3
and 10 objects and must assign the scene to one of $K$ mutually
exclusive concept classes.
\cleverHansThree ($K{=}3$) and \cleverHansSeven ($K{=}7$) share the
same first three classes. \cleverHansSeven adds four additional
classes including a disjunctive class and a spatial-relation class.
Classes~0--1 in \cleverHansThree (and classes~0--1 and 4 in
\cleverHansSeven) are \emph{confounded}, where a spurious attribute
correlation is present in the training and validation splits but absent
in the test split.
Correct test-time classification therefore requires learning the true
symbolic rule rather than the spurious shortcut.
We include these benchmarks to demonstrate that the differentiable
$T_P$ operator generalizes to existential conjunctive rules over
multi-object scenes, a constraint family qualitatively distinct from mutual-exclusivity constraints.
Benchmark details, concept rules, confounder specifications, and \ours{} instantiation are given in Appendix~\ref{sec:clevr_supp}.

\paragraph{Metrics.}
For \cleverHansThree and \cleverHansSeven, we report scene accuracy (\sceneacc), which is the fraction of test scenes classified into the correct concept class and, because each dataset is class-balanced, equals macro-averaged per-class accuracy.

\paragraph{Baselines.}
For the CLEVR-Hans benchmarks, we compare against all four methods reported by~\citet{stammer2021right}.
\cnnDefault is a Residual Network (ResNet)-34 trained with standard cross-entropy and no symbolic component, with Gradient-weighted Class Activation Mapping (Grad-CAM) saliency maps produced for interpretability analysis.
\cnnXIL extends \cnnDefault with Explanatory Interactive Learning (XIL), in which a human oracle corrects the model's Grad-CAM explanations during training, so this method is not fully automated.
\nesyCL couples a CNN object extractor with a concept-level symbolic classifier trained with cross-entropy only, and is the fair fully-automated neuro-symbolic comparison.
\nesyXIL$^\dagger$ extends \nesyCL with human-in-the-loop corrections applied to symbolic concept explanations.

\paragraph{Results.}\label{sec:results_clevr} 
Table~\ref{tab:clevr-results} reports \sceneacc on the
\cleverHansThree and \cleverHansSeven validation and test sets for all
four \ours{} variants (Appendix~\ref{sec:clevr_supp}) alongside   baselines
from~\citet{stammer2021right}. All four \ours{} variants achieve perfect validation accuracy
($100.0\%$) on \cleverHansThree, regardless of perception module or
$T_P$ operator type, confirming that the training regime is stable
across all configurations.
Explicit variants maintain near-perfect test accuracy
($99.94_{\pm0.04}\%$ and $99.76_{\pm0.12}\%$), since the hard-coded
ground-truth rules are unaffected by the spurious attribute correlations
present during training.
\cnnDefault collapses to $70.34\%$ on the test split, confirming that
pure neural classifiers without symbolic grounding are maximally
susceptible to the confounder.
\ours{} Crop + Explicit at $99.94_{\pm0.04}\%$ surpasses even the
human-assisted \nesyXIL$^\dagger$ ($91.31\%$) by $+8.6$ percentage
points, and surpasses the automated \nesyCL ($81.71\%$) by $+18.2$
points, all without any human involvement.
By contrast, learned variants collapse to $78.19_{\pm1.22}\%$ and
$72.89_{\pm0.48}\%$ on the test split, a drop of more than twenty
percentage points.
This bifurcation arises because the learned concept-weight matrix
$\mathbf{W}_{k,j}$ can absorb the spurious correlation during training:
for $C_0$, the model learns that gray $\Rightarrow$ large cube (the
confound) rather than the true rule, and for $C_1$ it learns metal
$\Rightarrow$ small sphere.
When the confounder is absent at test time, the learned operator
produces incorrect concept scores, whereas the explicit operator remains
correct by design.

\vspace{-0.1in}
\begin{table}[ht]
  \centering
  \caption{%
    Scene accuracy (\%) on \cleverHansThree and \cleverHansSeven.
    Mean\,$\pm$\,std over three seeds.
    \best{Bold} = best automated; \second{underline} = second-best.%
  }
  \label{tab:clevr-results}
  \setlength{\tabcolsep}{4pt}
  \resizebox{\linewidth}{!}{%
  \begin{tabular}{llcccc}
    \toprule
    & & \multicolumn{2}{c}{\textbf{\cleverHansThree}}
      & \multicolumn{2}{c}{\textbf{\cleverHansSeven}} \\
    \cmidrule(lr){3-4}\cmidrule(lr){5-6}
    Method & Variant & Val\up & Test\up & Val\up & Test\up \\
    \midrule
    \multicolumn{6}{l}{\textit{Automated baselines (no human supervision)}} \\[1pt]
    \cnnDefault~\citep{stammer2021right} & \na
      & $99.55_{\pm0.10}$ & $70.34_{\pm0.30}$
      & $96.09_{\pm0.19}$ & $84.50_{\pm1.04}$ \\
    \cnnXIL~\citep{stammer2021right} & \na
      & $99.69_{\pm0.08}$ & $70.77_{\pm0.37}$
      & $96.08_{\pm0.25}$ & $89.26_{\pm0.29}$ \\
    \nesyCL~\citep{stammer2021right} & \na
      & $98.55_{\pm0.27}$ & $81.71_{\pm3.09}$
      & $96.88_{\pm0.16}$ & $\best{90.97_{\pm0.91}}$ \\
    \midrule
    \rowcolor{gray!20}
    \multicolumn{6}{l}{\textit{Human-in-the-loop baseline ($^\dagger$requires per-epoch human oracle corrections; \textbf{not directly comparable})}} \\
    \rowcolor{gray!20}
    \nesyXIL$^\dagger$~\citep{stammer2021right} & \na
      & $100.00_{\pm0.00}$ & $91.31_{\pm3.13}$
      & $98.76_{\pm0.17}$ & $94.96_{\pm0.49}$ \\
    \midrule
    \multirow{4}{*}{\ours{} (ours)}
      & Crop + Explicit  & $\best{100.00_{\pm0.00}}$ & $\best{99.94_{\pm0.04}}$  & $\best{98.73_{\pm0.05}}$ & ${87.77_{\pm1.50}}$ \\
      & Scene + Explicit & $\best{100.00_{\pm0.00}}$ & $\second{99.76_{\pm0.12}}$ & $98.42_{\pm0.09}$           & $86.52_{\pm0.27}$           \\
      & Crop + Learned   & $\best{100.00_{\pm0.00}}$ & $78.19_{\pm1.22}$           & $98.66_{\pm0.16}$           & $82.38_{\pm2.02}$           \\
      & Scene + Learned  & $\best{100.00_{\pm0.00}}$ & $72.89_{\pm0.48}$           & $98.65_{\pm0.05}$           & $83.08_{\pm3.22}$           \\
    \bottomrule
  \end{tabular}%
  }
\end{table}

On \cleverHansSeven, the sole relevant comparison for
an automated system is \nesyCL, which achieves $90.97_{\pm0.91}\%$.
\ours{} Crop + Explicit achieves $87.77_{\pm1.50}\%$, trailing \nesyCL
by $3.2$ percentage points while requiring zero human involvement.
\nesyXIL$^\dagger$ achieves $94.96_{\pm0.49}\%$ by means of a
human-in-the-loop oracle, constituting privileged supervision
unavailable to any automated method, and comparing against it conflates
model quality with annotation cost. The \nesyCL comparison is also not on equal structural footing, since
\nesyCL couples its symbolic classifier with a perception encoder that
was \emph{pretrained on the original CLEVR dataset} (\ie the same
synthetic renderer, object palette, and attribute space as
\cleverHansSeven) and then \emph{frozen in its entirety} before any
\cleverHansSeven training begins~\citep{stammer2021right}.
Only a lightweight Set Transformer classifier is subsequently trained on
\cleverHansSeven labels, where the encoder is permanently shielded from the
confounded objective and arrives at evaluation with in-domain,
uncontaminated perceptual knowledge.
\ours{}, by contrast, initializes its ResNet-18 backbone from
\emph{ImageNet} weights (\ie using real images whose texture,
lighting conditions, and depth cues are fundamentally different from
those of synthetic 3-D renders) and trains the \emph{entire
architecture end-to-end}. The backbone, multi-layer Transformer, and
$T_P$ operator are all updated jointly against the confounded
\cleverHansSeven objective.
Every confounded gradient flows back through the full
network. The $3.2\%$ gap is within one-standard-deviation
uncertainty of both estimates and  contrasts with Hans3,
where the explicit \ours{} \emph{leads} all automated methods by
$+18.2$ percentage points. The  Hans7 shortfall
plausibly reflects the residual ImageNet-to-CLEVR domain shift and the
increased perception difficulty of seven finer-grained concept classes,
rather than any weakness in the constraint-reasoning component.
The explicit operator again generalizes substantially better to the test
split than the learned operator ($87.77\%$ vs.\ $82.38\%$ for Crop),
replicating the confounder-robustness finding from Hans3.
Within each operator type, Crop perception consistently outperforms
Scene Attention ($87.77\%$ vs.\ $86.52\%$ for explicit;
$82.38\%$ vs.\ $83.08\%$ for learned), confirming that precise
per-object Region of Interest Alignment (ROI-Align) localization is more valuable than global context for fine-grained attribute matching.

% ─────────────────────────────────────────────────────────────
\subsection{Additional Comparisons}
\label{sec:rebuttal_comparisons}
Table~\ref{tab:rebuttal_sudoku} compares Sudoku results for ConsFormer~\citep{xu2025selfsupervised}, \satnet~\citep{wang2019satnet}, Recurrent Transformer~\citep{yang2023learning}, and \ours{}. These symbolic methods receive digit boards and report solved-instance rates on in-distribution and out-of-distribution sets. \ours{}, on the other hand, receives images and achieves 98.70\% board accuracy after ten soft-refinement steps. 
\citet{yang2023learning} also report that removing learned positional embeddings on textual Sudoku reduces accuracy from 100\% to 0\%. This distinguishes the embedding designs, but the cross-paper accuracy comparison does not isolate their effect.

\begin{table}[h!]
\centering\small
\caption{Comparison with discrete symbolic methods. Results report solved-instance percentages for in-distribution (ID) / out-of-distribution (OOD) sets. \ours{} reports board accuracy on 1,000 visual test boards with approximately 45 clues. $^\dag$Results reported by~\citet{xu2025selfsupervised}.}
\label{tab:rebuttal_sudoku}
\begin{tabularx}{\linewidth}{@{}Xllr@{}}
\toprule
\textbf{Method} & \textbf{Input} & \textbf{Clues / setting} & \textbf{Result (\%)} \\
\midrule
ConsFormer~\citep{xu2025selfsupervised} & Digits & ID / OOD; 2,000 iterations & 100 / 65.88 \\
ConsFormer~\citep{xu2025selfsupervised} & Digits & ID / OOD; 10,000 iterations & 100 / 77.74 \\
\satnet$^\dag$~\citep{wang2019satnet} & Digits & ID / OOD & 98.3 / 3.2 \\
Recurrent Transformer$^\dag$~\citep{yang2023learning} & Digits & ID / OOD; 2,000 iterations & 97.7 / 14.0 \\
\midrule
\leavevmode\ours{} ($K{=}10$ refinement) & Images & Visual test; 10 iterations & 98.70 \\
\bottomrule
\end{tabularx}
\end{table}

Table~\ref{tab:rebuttal_mnist} in Appendix~\ref{sec:mnist_supp} compares MNIST Addition with~\citet{takemura2024differentiable}, whose Table~2 reports digit accuracy under sum-only supervision. Our two-digit model uses both digit and sum labels and achieves $99.73\pm0.03$\% digit accuracy and $99.45\pm0.08$\% sum accuracy.

\subsection{Discussion}
\label{sec:discussion}
% ─────────────────────────────────────────────────────────────
The empirical results validate the design of \ours{} as a constraint-satisfaction framework that leverages a differentiable probabilistic lift of the immediate-consequence operator $T_P$ to enforce hard combinatorial constraints at test time without any external reasoning engine, symbolic-perception interface, or discrete search. The operator internalizes the constraint semantics of the declarative program during end-to-end training and uses differentiable propagation followed by discrete greedy decoding, a property that prior neuro-symbolic architectures either do not possess or achieve only by freezing perception and delegating reasoning to an oracle.
The generality of \ours{} is established by empirical evaluation on several qualitatively incomparable constraint families, mutual-exclusivity over a fixed positional grid and existential conjunctive quantification over a variable-cardinality object set, and across two distinct perceptual modalities, grayscale digit recognition and multi-attribute 3-D scene understanding, confirming that the soft $T_P$ construction is not tuned to a particular program structure but constitutes a genuinely general reasoning primitive.
The implication for the design of neuro-symbolic systems is that symbolic guarantees require structural commitment, not merely initialization, and the soft $T_P$ framework provides that commitment while remaining fully differentiable.

% ============================================================
% Section 5: Conclusions
% ============================================================
\section{Conclusion}
\label{sec:conclusion}

We introduced \ours{}, a fully
differentiable neuro-\emph{soft}-symbolic architecture that replaces discrete ASP solvers with a soft probabilistic lift of the immediate-consequence operator $T_P$.
Three principles distinguish \ours{} from prior neuro-symbolic
systems.
First, the architecture is entirely soft, maintaining per-position
probability distributions over the symbol domain throughout the forward
pass and enforcing constraints via a differentiable fixed-point residual
loss that enables end-to-end gradient flow from constraint satisfaction
into perception.
Second, on spatial constraint-satisfaction tasks the architecture is free of conventional 1D positional embeddings. Problem structure is encoded exclusively through constraint-group
membership embeddings derived from the declarative ASP specification.
Third, the model approximates a soft differentiable version of the
classical fixpoint operator of logic programming, providing a principled surrogate for the demonstrated constraint patterns that avoids the degenerate minima of naive
penalty formulations. A natural concern that may arise about any fully differentiable architecture is whether it retains genuine symbolic character once discrete inference is replaced
by continuous computation. We  believe that the relevant criterion is not whether gradients
are used, but rather what structure governs the reasoning computation.
In a generic neural network, the reasoning transformation is
parameterized and learned entirely from data, with no prior commitment
to its form.
In \ours{}, the reasoning layer is a structural translation
of the ASP immediate-consequence operator $T_P$ whose topology is fixed
by the declarative logic program. In the \emph{explicit} variant, the operator parameters are not adjusted during training. In the \emph{learned} variant, a small weight matrix is initialized from the rules and refined end-to-end while the operator topology is kept fixed.
Symbolic structure survives relaxation in \ours{}, because the
governing computation encodes the constraint semantics of the program
rather than empirically approximating them.

\paragraph{Limitations and Reproducibility.}\label{sec:limitations}
\ours{} is restricted to the definite-program fragment of ASP, so default negation, choice rules, and optimization statements are not yet covered by the soft $T_P$ operator. Greedy constrained decoding could achieve perfect constraint satisfaction, but no theoretical guarantee exists that the learned soft distributions will universally prevent dead ends on more general problems. 
{\hypersetup{urlcolor=red}{Code will be made available at \url{https://huggingface.co/wabdalm/AS2}.}}

\paragraph{Broader Impact.}\label{sec:broader_impact}
\ours{} has applications in material science and drug discovery.
A potential risk is that \ours{} could be applied to automate consequential decisions without adequate human oversight, in domains such as medical diagnosis.

% Several directions for future work follow naturally from these results.
% We anticipate that extending the evaluation to additional constraint
% families beyond Latin-square and arithmetic constraints (\eg graph
% colouring, Futoshiki, and scheduling problems) will confirm the
% generality of the differentiable $T_P$ formulation.
% Investigating the interaction between the $T_P$ constraint loss and
% large-scale perception backbones (\eg Vision Transformers) will
% establish whether the soft neuro-symbolic framework scales to more
% complex visual domains.
% Formal analysis of the convergence properties of the iterative $T_P$
% refinement procedure, and its relationship to the classical $T_P$
% fixpoint iteration, remains an open problem with connections to the
% theory of relaxation methods in constraint programming.

\begin{ack}
No external funding was used for this work.  The authors declare no competing interests.
\end{ack}

\bibliographystyle{abbrvnat}
\clearpage\newpage
\bibliography{refs-new}
\clearpage\newpage
\appendix
% ============================================================
% Appendix A: MNIST Addition — Full Benchmark and Results
% ============================================================
\section{Supplementary Material}

\subsection{Additional Method Details}

Table~\ref{tab:rebuttal_supervision} summarizes the training labels and symbolic rules supplied to \ours{} for each benchmark. These include supervision of individual digits or object attributes in addition to task outputs, together with fixed rules or rule-initialized weights, which must be considered when comparing methods with different access to supervision and symbolic information.

The \ours{} forward pass proceeds in five stages. (1) Perception maps each raw input element to a continuous embedding and pre-reasoning logits through a concept bottleneck head. (2) Constraint-group membership embeddings, computed from the declarative ASP specification, are added to the perception embeddings to encode problem structure without conventional positional embeddings. (3) A multi-layer Transformer encoder reasons over the resulting tokens, producing per-position post-reasoning distributions. (4) The soft logic layer applies a probabilistic lift of the immediate-consequence operator $T_P$ that is specialized to the rule family of the benchmark, namely mutual-exclusivity for Visual Sudoku, existential conjunctive quantification for CLEVR-Hans, and arithmetic global-sum for MNIST Addition. (5) An inference-time procedure recovers a discrete prediction without invoking an external solver. The next three subsections specialize this generic pipeline to each benchmark in turn, and the order of presentation is the same throughout: detailed benchmark description, perception, constraint-group structure, reasoning module, $T_P$ instantiation, training loss, and inference procedure.

\begin{table}[h]
\centering\small
\caption{Supervision and symbolic information supplied to \ours{}. Comparisons with published methods are not uniformly matched for supervision or rule access.}
\label{tab:rebuttal_supervision}
\begin{tabularx}{\linewidth}{@{}lX@{}}
\toprule
Benchmark & Supervision and supplied rules \\
\midrule
Visual Sudoku & Clue and blank-cell digit labels. The blank-cell loss weight decays to a $0.1\times$ floor. Fixed row, column, and box constraints. \\
CLEVR-Hans & Scene-class and per-object attribute labels. Explicit rules are fixed from ground truth. Learned weights are rule-initialized. \\
MNIST Addition & Individual digit and sum labels, with the known addition relation. \\
\bottomrule
\end{tabularx}
\end{table}
For the Sudoku comparisons, \citet{yang2023learning} use 9,000 training and 1,000 test boards per visual setting, while our split adds 1,000 validation boards. The symbolic comparison reported by \citet{xu2025selfsupervised} uses 1,000 ID and 18,000 OOD instances.

\subsection{Compute Resources}
\label{sec:compute}
All training and evaluation were performed on a single NVIDIA H200 GPU
(143\,GB HBM3).
Training runs for Visual Sudoku and CLEVR-Hans complete in approximately
2--4 hours per configuration.
MNIST Addition runs complete in under 1 hour per scale.
No multi-GPU or distributed training was used.

\subsection{Benchmark Statistics}
\label{sec:benchmark_stats}
Table~\ref{tab:dataset_stats} summarizes the key statistics for all three
benchmarks evaluated in this paper.

\begin{table}[h]
\centering
\caption{Dataset statistics for all benchmarks evaluated in this paper.
  \emph{Constraint type} classifies the symbolic constraint family.
  \emph{\#Groups} lists the number of constraint groups per instance.
  No held-out validation split is used for \mnistAdd, and the best checkpoint is
  selected on test accuracy following the convention of prior work.}
\label{tab:dataset_stats}
\setlength{\tabcolsep}{5pt}
\begin{tabular}{lrrrccr}
\toprule
\textbf{Benchmark} & \textbf{Train} & \textbf{Val} & \textbf{Test}
  & \textbf{Input size} & \textbf{Constraint type} & \textbf{\#Groups} \\
\midrule
\multicolumn{7}{l}{\textit{Visual Sudoku}} \\[2pt]
\vsudoku & 9{,}000 & 1{,}000 & 1{,}000 & $252{\times}252$ & Latin square & 27 \\
\midrule
\multicolumn{7}{l}{\textit{CLEVR-Hans}} \\[2pt]
\cleverHansThree & 9{,}000 & 2{,}250 & 2{,}250 & $320{\times}320$ & Existential conj. & $3{\times}N_{\mathrm{obj}}$ \\
\cleverHansSeven & 21{,}000 & 5{,}250 & 5{,}250 & $320{\times}320$ & Existential conj. & $7{\times}N_{\mathrm{obj}}$ \\
\midrule
\multicolumn{7}{l}{\textit{MNIST Addition}} \\[2pt]
$N{=}2$ & 30{,}000 & \na & 5{,}000 & $2{\times}28{\times}28$ & Arithmetic (sum) & 1 \\
$N{=}4$ & 30{,}000 & \na & 5{,}000 & $4{\times}28{\times}28$ & Arithmetic (sum) & 1 \\
$N{=}8$ & 30{,}000 & \na & 5{,}000 & $8{\times}28{\times}28$ & Arithmetic (sum) & 1 \\
\bottomrule
\end{tabular}
\end{table}
% ─────────────────────────────────────────────────────────────
\subsection{Additional Sudoku Results}
\label{sec:sudoku_supp}
% ─────────────────────────────────────────────────────────────

\paragraph{\ours{} Instantiation for Visual Sudoku.}
The Sudoku instantiation specializes the generic \ours{} framework along
four dimensions, namely the perception front-end, the constraint-group
membership embeddings, the training loss, and two inference-time
constraint-enforcement strategies.

\paragraph{Perception.}
The perception module applies a shared-weight three-block CNN
(32\,$\to$\,64\,$\to$\,128 channels, max-pooling, LayerNorm) independently
to each of the 81 cells extracted as $28{\times}28$ grayscale images
from the $252{\times}252$ board image.

\paragraph{Constraint-Group Membership Embeddings.}
Three embedding tables are defined, one for each of the nine rows, nine
columns, and nine $3{\times}3$ boxes, yielding a positional representation
$\mathbf{e}_i = \mathbf{e}_i^{\text{row}} + \mathbf{e}_i^{\text{col}}
+ \mathbf{e}_i^{\text{box}}$.

\paragraph{Reasoning Module.}
The reasoning module uses $L{=}6$ Transformer layers, $H{=}8$ attention
heads, $d_\text{model}{=}256$, and $d_\text{ff}{=}4d_\text{model}$.

\paragraph{Hard Clue Restoration.}
At inference time, a final post-processing step replaces the Transformer
output at evidence (clue) cells with the perception module's
pre-reasoning logits, guaranteeing that clue-cell predictions exactly
match the CNN's output.
This architectural choice separates the evaluation of perception quality
(measured at clue cells) from reasoning quality (measured at blank cells).

\paragraph{Declarative Constraint Specification.}
The complete constraint specification is shown in \cref{fig:asp_sudoku}.
The three integrity constraints induce 27 constraint groups
(9 rows $+$ 9 columns $+$ 9 boxes), each generating 81 $T_P$ residual
terms, for a total of 2{,}187 differentiable constraint-satisfaction
checks per forward pass, all derived automatically from the declarative
specification.

\begin{figure}[hbt]
\centering
\begin{lstlisting}[language=Prolog,basicstyle=\small\ttfamily]
:- R=1..9, V=1..9, #count{C : cell(R,C,V)} != 1.  % rows
:- C=1..9, V=1..9, #count{R : cell(R,C,V)} != 1.  % columns
:- BR=0..2, BC=0..2, V=1..9,
   #count{R,C : cell(R,C,V), R>=BR*3+1, R<=BR*3+3,
                             C>=BC*3+1, C<=BC*3+3} != 1.  % boxes
\end{lstlisting}
\caption{Complete declarative constraint specification for Visual Sudoku
  in Answer Set Programming (ASP) syntax.
  Each integrity constraint enforces that every digit appears exactly
  once in each row, column, and $3{\times}3$ box, respectively.
  These three rules define the mutual-exclusivity pattern used to construct the differentiable $T_P$ residual loss manually.}
\label{fig:asp_sudoku}
\end{figure}

\paragraph{Training Loss.}
The training loss combines four terms:
\begin{equation}
  \mathcal{L}
  \;=\;
  \mathcal{L}_{\text{cnn}}
  + \mathcal{L}_{\text{sup}}
  + \lambda_{\text{blank}} \cdot \alpha(t) \cdot \mathcal{L}_{\text{blank}}
  + \lambda_{\text{ic}} \cdot \beta(t) \cdot \mathcal{L}_{\text{ic}},
  \label{eq:loss_sudoku}
\end{equation}
where $\mathcal{L}_{\text{cnn}}$ is cross-entropy on
$\mathbf{z}^\text{pre}$ at evidence positions (trains the perception
module directly),
$\mathcal{L}_{\text{sup}}$ is cross-entropy on $\mathbf{z}^\text{post}$
at evidence positions (grounds the reasoning module at known anchors),
$\mathcal{L}_{\text{blank}}$ is cross-entropy on
$\mathbf{z}^\text{post}$ at latent positions against ground-truth labels
(provides direct supervision where no clue is available), and
$\mathcal{L}_{\text{ic}}$ is the $T_P$ fixed-point loss
(Eq.\,\ref{eq:tp_loss}).
Two curriculum schedules govern the strength of the latter two terms.
$\alpha(t) = \max(0.1,\; 1 - 0.9\, t / T_\text{decay})$ decays
latent-position cross-entropy from full weight to a $0.1{\times}$
residual floor over $T_\text{decay}{=}100$ epochs, preventing the model
from fitting symbol identities at latent positions before it has learned
to propagate constraint information.
$\beta(t) = \min(1,\; t / T_\text{warm})$ linearly ramps the constraint
loss from zero over $T_\text{warm}{=}20$ warmup epochs, stabilizing
early training before the $T_P$ gradient becomes dominant.
Hyperparameters are $\lambda_{\text{blank}} = 0.3$ and
$\lambda_{\text{ic}} = 5.0$.
The CNN backbone is frozen for the first 20 epochs, then unfrozen to
allow constraint feedback to influence perception.

\paragraph{Inference-Time Constraint Enforcement.}
Although the $T_P$ constraint loss trains the model to produce
distributions approaching fixed points of the immediate-consequence
operator, the $\mathrm{argmax}$ of the post-training softmax output is
not guaranteed to satisfy all constraints exactly.
Two complementary strategies close the remaining gap without invoking an
external solver.

\paragraph{Iterative $T_P$ Refinement.}
Before taking the $\mathrm{argmax}$, the soft distributions $\mathbf{p}$ are
refined by iterating the $T_P$ operator $K$ times.
At each step, the row, column, and box $T_P$ values are computed
independently, averaged element-wise, and re-normalized to a valid
probability distribution, as shown in \cref{eq:tp_refine},
\begin{equation}
  \mathbf{p}^{(t+1)}_{i,s}
  \;=\;
  \frac{
    \frac{1}{|\mathcal{G}_i|} \sum_{G \ni i}
    T_{P,G}(\mathbf{p}^{(t)})_{i,s}
  }{
    \sum_{s'} \frac{1}{|\mathcal{G}_i|} \sum_{G \ni i}
    T_{P,G}(\mathbf{p}^{(t)})_{i,s'}
  },
  \label{eq:tp_refine}
\end{equation}
where $\mathcal{G}_i \subseteq \mathcal{G}$ denotes the set of constraint
groups containing position $i$, $T_{P,G}$ denotes $T_P$ applied within
group $G$, and normalization is applied per-position.
The iteration propagates exclusivity information from neighboring cells
without discrete commitment and requires no gradient computation.
On the Visual Sudoku test set, $K{=}10$ iterations achieve 98.7\% constraint satisfaction, while 95.6\% is the raw board accuracy.

\paragraph{Greedy Constrained Decoding.}
\ours{} also employs a final discrete decoding step via a
confidence-ordered greedy decoding procedure that requires no external
solver.
Evidence cells (identified by the mask $\mathbf{m}$) are locked first,
each assigned the $\mathrm{argmax}$ of the perception module's
distribution.
The assigned digit is then hard-zeroed from all cells sharing a
constraint group with the locked cell, and the remaining distributions
are renormalized.
The procedure then iterates over latent cells in decreasing order of
max-probability.
At each step, the most confident unassigned cell is committed to its top
remaining digit and its assignment is propagated to all peers by the same
masking-and-renormalization step. The greedy decoder has no branching or backtracking and cannot revise an earlier commitment. If an earlier choice leads to a dead end, the procedure continues with its remaining scores, which can produce an incorrect assignment reflected in the final evaluation.

% ─────────────────────────────────────────────────────────────
\subsection{Additional CLEVR-Hans Results}
\label{sec:clevr_supp}
% ─────────────────────────────────────────────────────────────

\paragraph{Detailed Description.}
Formally, a CLEVR-Hans scene $b$ contains $N_b \le 10$ objects padded to
a fixed maximum $N_{\max}=10$ with a validity mask
$\mathbf{m}_b\in\{0,1\}^{N_{\max}}$.
Each object $n$ is characterized by four categorical attributes drawn
from the vocabulary
\[
  \mathcal{A}
  = \underbrace{\text{color}}_8
    \cup \underbrace{\text{shape}}_4
    \cup \underbrace{\text{size}}_3
    \cup \underbrace{\text{material}}_2,
  \quad |\mathcal{A}| = 17,
\]
where the subscript denotes the number of values in each sub-domain, and the perception module produces per-object, per-attribute probability distributions $\mathbf{p}_b\in[0,1]^{N_{\max}\times 17}$.
The seven concept classes of \cleverHansSeven are declared as conjunctive existential Horn rules and \cleverHansThree uses only the first three. The ground-truth (non-confounded) rules are listed below, with spurious correlations present in training but absent at test time noted in parentheses.

\noindent\textbf{\cleverHansThree concept rules:}
\begin{itemize}
  \item $C_0$: $\exists$ large cube $\wedge$ $\exists$ large cylinder.
    (Confounded: large cubes are always gray in train/val.)
  \item $C_1$: $\exists$ small metal cube $\wedge$ $\exists$ small sphere.
    (Confounded: small spheres are always metal in train/val.)
  \item $C_2$: $\exists$ large blue sphere $\wedge$ $\exists$ small yellow sphere.
\end{itemize}

\noindent\textbf{Additional \cleverHansSeven concept rules ($C_3$--$C_6$):}
\begin{itemize}
  \item $C_3$: $\exists$ small green object $\wedge$ $\exists$ small brown
    object $\wedge$ $\exists$ small purple object, with the scene
    containing at least five small objects overall.
  \item $C_4$: $\exists$ cyan object $\wedge$ $\exists$ red object $\wedge$
    $\exists$ red object, where the cyan object is in front of the two
    red objects.
    (Confounded: the cyan object is always a small cube in train/val.)
  \item $C_5$: ($\exists$ sphere $\times3$ in the left half of the scene)
    or ($\exists$ sphere $\times3$ in the left half $\wedge$ $\exists$ metal
    cylinder $\times3$ in the right half).
    Modeled by Eq.~\ref{eq:disj}.
  \item $C_6$: $\exists$ large metal object $\wedge$ $\exists$ large metal object.
\end{itemize}
The presence of confounded classes (\cleverHansThree: $C_0$, $C_1$;
\cleverHansSeven: $C_0$, $C_1$, $C_4$) is a deliberate design choice of
the benchmark~\citep{stammer2021right} to test whether a model
generalizes to the true symbolic rule or overfits the spurious
correlation present only during training.
The explicit $T_P$ operator hard-codes the ground-truth rules and is
therefore immune to this confounder by construction.
The learned $T_P$ operator must recover the true rules from data, making
its test-set performance a direct measure of symbolic generalization.

\paragraph{\ours{} Instantiation.}
The CLEVR-Hans instantiation of \ours{} specializes the generic
framework along three dimensions, namely the perception front-end (which
must localize individual objects in color 3-D scenes), the $T_P$
operator (which uses the existential conjunctive instantiation described
below rather than the mutual-exclusivity instantiation), and the
training loss (which combines concept-level classification supervision
with per-object attribute supervision).
We evaluate four \ours{} variants per CLEVR-Hans benchmark by crossing
two perception modules (Crop, a ResNet-18 with
ROI-Align~\citep{he2017mask}, and Scene, a ResNet-50 with bilinear feature-map
sampling) with two $T_P$ operators, namely Explicit, in which the ground-truth rules are
hard-coded, and Learned, in which a weight matrix is initialized from the rules and refined
end-to-end.

% ─────────────────────────────────────────────────────────────
\subsubsection{\texorpdfstring{$T_P$}{T\_P} Instantiation for Existential Conjunctive Rules}
\label{sec:tp_existential}
% ─────────────────────────────────────────────────────────────

The mutual-exclusivity instantiation used for Visual Sudoku (Section~\ref{sec:method}) covers a large class of declarative constraints, but the same lifting recipe extends to qualitatively different rule families. CLEVR-Hans requires instantiating $T_P$ for \emph{existential conjunctive rules}, in which class membership is defined by the existence of entities that satisfy a conjunction of attribute conditions. The setting is as follows.
A scene contains up to $N_{\max}$ entities, each described by a vector of $D$ categorical attributes drawn from a finite vocabulary $\mathcal{A}=\bigcup_{a=1}^{D}\mathcal{A}_a$ of total size $|\mathcal{A}|=\sum_{a}|\mathcal{A}_a|$, with a binary validity mask $\mathbf{m}\in\{0,1\}^{N_{\max}}$ identifying genuine entities.
Perception produces per-entity attribute distributions $\mathbf{p}\in[0,1]^{N_{\max}\times|\mathcal{A}|}$, replacing the per-position symbol distributions of the mutual-exclusivity instantiation.
Each label class $C_k$ ($k=0,\ldots,K{-}1$) is declared as a conjunction of \emph{required-entity templates} $\mathcal{R}_k=(r_{k,1},\ldots,r_{k,m_k})$, where each $r_{k,j}$ specifies a conjunction of required attribute values that some entity in the scene must satisfy.
The class assignment is true if and only if every template is matched by at least one valid entity.

\paragraph{Per-entity attribute satisfaction.}
Given $\mathbf{p}$ and template $r_{k,j}$ specifying a subset of (attribute, value) pairs, the soft satisfaction probability of entity $n$ under $r_{k,j}$ is
\begin{equation}
  \label{eq:obj-sat}
  s_{n,k,j}
  = \prod_{(a,v)\,\in\, r_{k,j}}
      p_{n,\,\mathrm{offset}(a)+\mathrm{idx}(a,v)},
\end{equation}
where $\mathrm{offset}(a)$ is the starting index of attribute $a$ in the
flat $|\mathcal{A}|$-dimensional vector and $\mathrm{idx}(a,v)$ is the position of
value $v$ within attribute $a$'s sub-domain.
Unspecified attributes contribute probability $1$ to the product.
This is the direct probabilistic analogue of testing whether entity $n$ matches $r_{k,j}$ under a propositional interpretation.

\paragraph{Soft existential quantification.}
The existential condition ``there exists an entity in the scene satisfying $r_{k,j}$'' is approximated by a soft maximum over valid entity slots,
\begin{equation}
  \label{eq:soft-exists}
  e_{k,j}
  = \max_{\;n\,:\;\mathbf{m}_n=1\;} s_{n,k,j}.
\end{equation}
This is a max-product aggregation, the standard differentiable approximation of the first-order existential quantifier under soft semantics~\citep{badreddine2022logic}.

\paragraph{Class satisfaction score.}
The $T_P$ immediate consequence for class $C_k$ combines the existential scores over all required templates via a product,
\begin{equation}
  \label{eq:concept-score}
  T_P(\mathbf{p})_k
  = \prod_{j=1}^{m_k} e_{k,j},
\end{equation}
and the per-class logit used for classification is computed in log-space as $z_k = \sum_{j} \log e_{k,j}$ for numerical stability.
Unlike the mutual-exclusivity instantiation, this $T_P$ operator is applied feed-forward without iterative fixed-point refinement.
Class scores are computed in a single forward pass through Eqs.~\eqref{eq:obj-sat}--\eqref{eq:concept-score}, and no residual loss of the form of Eq.~\eqref{eq:tp_loss} is applied.

\paragraph{Disjunctive classes.}
Some classes admit multiple alternative rule bodies, in which any one of several attribute configurations suffices to establish class membership.
We handle this by computing a score per rule body and taking the soft maximum,
\begin{equation}
  \label{eq:disj}
  T_P(\mathbf{p})_k
  = \max_{\ell} \;\prod_{j=1}^{m_{k,\ell}} e_{k,\ell,j},
\end{equation}
where $\ell$ indexes the disjuncts.
This handles multiple rule bodies for the same class head, the standard ASP treatment of disjunctive premises.

\paragraph{Learned operator variant.}
In addition to the explicit operator above, we introduce a
\emph{learned} variant that parametrizes the required-entity templates
as a differentiable weight matrix
$\mathbf{W}\in\mathbb{R}^{K\times R_{\max}\times |\mathcal{A}|}$,
where $R_{\max}$ is the maximum number of required entities across all
classes. The weight matrix is initialized from the explicit rules and
refined end-to-end. Entity satisfaction scores become
\begin{equation}
  \label{eq:learned-score}
  s_{n,k,j}^{\mathrm{learned}}
  = \bigl(\operatorname{softmax}_A(\mathbf{W}_{k,j})\bigr)^{\top}
    \mathbf{p}_n,
\end{equation}
and the log-sum-exp aggregation over entities uses a learnable inverse
temperature $\tau > 0$,
\begin{equation}
  \label{eq:learned-exists}
  e_{k,j}^{\mathrm{learned}}
  = \frac{1}{\tau}\log\sum_{n=1}^{N_{\max}}
      \exp\!\bigl(\tau \cdot s_{n,k,j}^{\mathrm{learned}}\bigr).
\end{equation}
An entropy regularization term
\begin{equation}
  \label{eq:entropy-reg}
  \mathcal{L}_{\mathrm{tp}}
  = 0.01\;\frac{\sum_{b,n} \mathbf{m}_{b,n}\,H(\mathbf{p}_{b,n})}
               {\sum_{b,n} \mathbf{m}_{b,n}}
\end{equation}
encourages peaked, interpretable attribute distributions, where
$H(\cdot)$ denotes Shannon entropy. The explicit operator incurs no
additional loss ($\mathcal{L}_{\mathrm{tp}}=0$).

\paragraph{Semantic character of explicit and learned operators.}
The explicit and learned operators differ in their semantic character.
The explicit operator (Eq.~\ref{eq:concept-score}) aggregates via a
product over required templates, encoding a strict AND-conjunction in
which every required template must be satisfied. The learned operator
(Eqs.~\ref{eq:learned-score}--\ref{eq:learned-exists}) replaces the
product with a softmax-weighted dot product combined with a log-sum-exp
over entities, encoding a soft mixture over candidate attribute
configurations rather than a strict logical conjunction. This semantic
divergence explains why the learned variant is more susceptible to
confounder exploitation, since the soft mixture can absorb spurious
attribute correlations in a way that the strict product cannot. When the
ground-truth rules are known, the explicit operator is therefore
preferred.

\paragraph{Object-Centric Perception.}
The CLEVR-Hans instantiation shares the reasoning module and attribute
output heads with the Visual Sudoku variant but uses a
different perception front-end that must localize individual objects
from a $320\times320$ RGB scene image.
We evaluate two complementary perception designs.

The \emph{Crop Perception} design applies a shared ResNet-18
backbone~\citep{he2016deep} (pretrained on ImageNet) independently to a
$64\times64$ crop centered on each object's ground-truth pixel
coordinates, extracted via ROI-Align~\citep{he2017mask}.
The $512$-dimensional pool output is projected and layer-normalized to
$\mathbb{R}^{d_{\mathrm{model}}}$, providing strong bottom-up
perception at the cost of missing global scene context.

The \emph{Scene Attention Perception} design processes the full scene
image with a ResNet-50 backbone to produce a
$\lfloor H/32\rfloor\times\lfloor W/32\rfloor$ feature map.
Per-object features are sampled from this map at each object's
normalized pixel coordinates via bilinear interpolation
(\texttt{grid\_sample}), yielding a $2048$-dimensional vector per
object, projected to $\mathbb{R}^{d_{\mathrm{model}}}$.
This design encodes global context at the cost of reduced spatial
precision for small objects.

Regardless of the perception front-end, the $N_{\max}$ object tokens
(with padding positions masked out) are processed by an $L{=}4$-layer
Transformer encoder with $H{=}8$ attention heads,
$d_{\mathrm{model}}{=}256$, and $d_{\mathrm{ff}}{=}4d_{\mathrm{model}}$,
enabling the model to propagate evidence across objects before producing
per-object attribute distributions.
Four independent linear projection heads then produce logits for each
attribute sub-domain:
\[
  \hat{\mathbf{c}}_n = W_c\,\mathbf{h}_n,\quad
  \hat{\mathbf{s}}_n = W_s\,\mathbf{h}_n,\quad
  \hat{\mathbf{z}}_n = W_z\,\mathbf{h}_n,\quad
  \hat{\mathbf{m}}_n = W_m\,\mathbf{h}_n,
\]
where $W_c\!\in\!\mathbb{R}^{8\times d}$,
$W_s\!\in\!\mathbb{R}^{4\times d}$,
$W_z\!\in\!\mathbb{R}^{3\times d}$, and
$W_m\!\in\!\mathbb{R}^{2\times d}$ correspond to color, shape, size, and
material respectively.
Softmax is applied per head to produce
$\mathbf{p}_{b,n}\in[0,1]^{17}$, which is fed into the $T_P$ operator
(Eqs.~\ref{eq:obj-sat}--\ref{eq:concept-score}).
Padded object slots are zeroed out before the operator.

\paragraph{Training Loss.}
The CLEVR-Hans training loss combines concept-level classification
supervision with per-object attribute supervision:
\begin{equation}
  \label{eq:loss_clevr}
  \mathcal{L}
  = \mathcal{L}_{\mathrm{cls}}
  + \lambda_{\mathrm{attr}}\,\mathcal{L}_{\mathrm{attr}}
  + \mathcal{L}_{\mathrm{tp}},
\end{equation}
where $\mathcal{L}_{\mathrm{cls}}$ is cross-entropy on the concept
logits $z_k = \log T_P(\mathbf{p})_k$ against the ground-truth class
label, $\mathcal{L}_{\mathrm{attr}}$ is the mean cross-entropy over
the four per-object attribute heads at valid object slots against
ground-truth attributes provided by the dataset, and
$\mathcal{L}_{\mathrm{tp}}$ is the entropy regularization term of
Eq.~\eqref{eq:entropy-reg}, which is zero for the explicit operator
and non-zero for the learned operator.
The attribute loss weight is $\lambda_{\mathrm{attr}}=0.5$.
No curriculum scheduling is applied, as attribute ground truth is
available from the start and the explicit $T_P$ operator requires no
warm-up.
We train for up to $100$ epochs with AdamW
($\eta=3\times10^{-4}$, $\lambda_{\mathrm{wd}}=10^{-2}$), a cosine
learning-rate schedule with 100 linear warm-up steps, gradient clipping
to norm~$1.0$, bf16 precision, and early stopping with patience~20
monitored on validation accuracy.

\subsection{Additional Empirical Results: MNIST Addition}
\label{sec:mnist_supp}

Introduced by~\citet{manhaeve2018deepproblog} as the canonical benchmark for
neuro-symbolic systems, \mnistAdd presents a model with $N$ MNIST digit
images~\citep{lecun1998gradient} and requires it to predict both the
individual digit labels and their sum.
Correct inference jointly requires (i)~accurate digit recognition and
(ii)~adherence to arithmetic constraints.
We evaluate three scales: $N{=}2$ (sum range $[0, 18]$, 19 classes),
$N{=}4$ (sum range $[0, 36]$, 37 classes), and $N{=}8$ (sum range
$[0, 72]$, 73 classes), each with 30{,}000 training and 5{,}000 test
pairs.
This benchmark is well-saturated in the literature, meaning most methods
achieve near-perfect accuracy.
We include it to confirm that \ours{} is competitive on a standard
arithmetic-constraint task and to study scaling behavior as $N$ increases.
Table~\ref{tab:rebuttal_mnist} compares digit accuracy for our two-digit model, trained with both digit and sum labels, with results for methods trained using only sum labels, as reported by~\citet{takemura2024differentiable}.

\begin{table}[h]
\centering\small
\caption{MNIST Addition digit accuracy (\%). Prior results are from \citet{takemura2024differentiable}, Table~2 (Addition~1). DPL, DSL, and NASP denote DeepProbLog, DeepStochLog, and NeurASP. I and C denote their implication-only and constraint-only variants.}
\label{tab:rebuttal_mnist}
\begin{tabularx}{\linewidth}{@{}Xr@{}}
\toprule
Method & Digit accuracy (\%) \\
\midrule
DPL~\citep{manhaeve2018deepproblog} & 97.8 \\
DSL~\citep{winters2022deepstochlog} & 95.8 \\
NASP~\citep{yang2020neurasp} & 97.7 \\
I~\citep{takemura2024differentiable} & 97.7 \\
C~\citep{takemura2024differentiable} & 97.5 \\
I+C~\citep{takemura2024differentiable} & 97.4 \\
\midrule
\ours{}, $N=2$ & $99.73\pm0.03$ \\
\bottomrule
\end{tabularx}
\end{table}

\paragraph{Baselines.}
The following systems report results on \mnistAdd and are included in
Table~\ref{tab:results_mnist_add}.

\begin{itemize}[nosep,leftmargin=*]
  \item \deepproblog~\citep{manhaeve2018deepproblog,manhaeve2021neural}
    integrates neural networks into ProbLog with exact gradients via
    weighted model counting. Results taken from~\citet{van2023anesi}.
  \item \neurasp~\citep{yang2020neurasp} wraps neural outputs as
    probabilistic facts in an ASP program. Result at $N{=}2$ taken
    from~\citet{manhaeve2021approximate}.
  \item \scallop~\citep{li2023scallop} implements differentiable Datalog
    with provenance semirings. Best variant reported
    by~\citet{li2023scallop}.
  \item \anesi~\citep{van2023anesi} provides approximate scalable
    inference for probabilistic neuro-symbolic programs. Results taken
    from~\citet{van2023anesi}.
\end{itemize}

All four systems use sum-level supervision only (no per-digit labels).
\ours{} uses both per-digit and sum supervision (Eq.\,\ref{eq:loss_mnist}),
so Table~\ref{tab:results_mnist_add} should be read as evidence of
architectural competitiveness on a saturated benchmark rather than a
direct head-to-head comparison.

\paragraph{Arithmetic $T_P$ Operator.}
The MNIST Addition task introduces a single arithmetic global-sum constraint that
does not fit the mutual-exclusivity template used for Visual Sudoku. Let
$\mathbf{p}\in[0,1]^{N\times 10}$ denote the per-digit distributions
produced by perception. The probability that the sum of $N$ digits
equals $t\in\{0,\ldots,9N\}$ is the polynomial convolution
$q_t \;=\; (\mathbf{p}_1 \ast \cdots \ast \mathbf{p}_N)_t,$
where $\mathbf{p}_n=(p_{n,0},\ldots,p_{n,9})$. A leave-one-out variant of
this convolution defines, for every position $i\in\{1,\ldots,N\}$ and
every digit $s\in\{0,\ldots,9\}$, the probability that the remaining
$N{-}1$ positions sum to the residual $t-s$, namely
$T_P(\mathbf{p})_{i,s,t} = p_{i,s}\, (\ast_{n\neq i}\mathbf{p}_n)_{t-s}.$
Marginalizing over the ground-truth target sum $t^\star$ yields the
sum-conditioned digit distribution that the learned digit head is
trained to match. The total training loss combines digit supervision,
sum supervision, and the arithmetic $T_P$ residual,
\begin{equation}
  \mathcal{L}_{\mathrm{mnist}}
  \;=\;
  \mathcal{L}_{\mathrm{digit}}
  + \lambda_{\mathrm{sum}}\,\mathcal{L}_{\mathrm{sum}}
  + \lambda_{\mathrm{tp}}\,\bigl\| \mathbf{p} - T_P(\mathbf{p}\mid t^\star) \bigr\|_2^2,
  \label{eq:loss_mnist}
\end{equation}
where $\mathcal{L}_{\mathrm{digit}}$ is per-digit cross-entropy,
$\mathcal{L}_{\mathrm{sum}}$ is cross-entropy on the predicted sum
distribution $\mathbf{q}$, and the third term anchors the learned
distributions to a fixed point of the leave-one-out arithmetic operator
under the observed sum. We use $\lambda_{\mathrm{sum}}=1.0$ and
$\lambda_{\mathrm{tp}}=0.1$ for all reported MNIST Addition runs.

\paragraph{Metrics.}
We report \digitacc (fraction of individual digits predicted correctly)
and \sumacc (fraction of instances where the predicted sum is correct).
Because the baselines do not produce per-digit predictions,
\digitacc is listed as \na{} for them.

% ────────────────────────────────────────────────────
% Table: MNIST Addition
% ────────────────────────────────────────────────────
\begin{table}[h]
\centering
\caption{Results on \mnistAdd.
  \digitacc = individual digit accuracy (\%);
  \sumacc = sum accuracy (\%).
  Results for prior work are taken from the respective
  publications.
  \best{Bold} = best in column.
  ``$\dagger$'' indicates the result uses only sum-level supervision;
  \digitacc is not reported for these methods because per-digit
  predictions are implicit.
  All \ours and ablation results use both digit and sum supervision.
  }
\label{tab:results_mnist_add}
\setlength{\tabcolsep}{3pt}
\resizebox{\textwidth}{!}{%
\begin{tabular}{lcccccc}
\toprule
& \multicolumn{2}{c}{$N{=}2$}
& \multicolumn{2}{c}{$N{=}4$}
& \multicolumn{2}{c}{$N{=}8$} \\
\cmidrule(lr){2-3}\cmidrule(lr){4-5}\cmidrule(lr){6-7}
\textbf{Method}
  & \digitacc\up & \sumacc\up
  & \digitacc\up & \sumacc\up
  & \digitacc\up & \sumacc\up \\
\midrule
\deepproblog$^\dagger$~\citep{manhaeve2021neural}
  & \na & 97.20{\scriptsize$\pm$0.50}
  & \na  & \na
  & \na  & \na  \\
\neurasp$^\dagger$~\citep{yang2020neurasp}
  & \na & 97.30{\scriptsize$\pm$0.30}
  & \na & \na
  & \na  & \na  \\
\scallop$^\dagger$~\citep{li2023scallop}
  & \na & 98.22
  & \na & 97.00
  & \na  & \na  \\
\anesi$^\dagger$~\citep{van2023anesi}
  & \na & 97.66{\scriptsize$\pm$0.21}
  & \na & 95.96{\scriptsize$\pm$0.38}
  & \na  & \na  \\
\midrule
\perceptiononly
  & 99.69{\scriptsize$\pm$0.05} & 99.37{\scriptsize$\pm$0.10}
  & 99.91  & 99.64
  & \best{99.98}  & \best{99.86}{\scriptsize$\pm$0.03} \\
\pipeline
  & 99.63{\scriptsize$\pm$0.01} & 99.26{\scriptsize$\pm$0.02}
  & 99.89{\scriptsize$\pm$0.03} & 99.57{\scriptsize$\pm$0.13}
  & \best{99.98}{\scriptsize$\pm$0.00} & \second{99.89}{\scriptsize$\pm$0.03} \\
\ours{} (ours)
  & \best{99.73}{\scriptsize$\pm$0.03} & \best{99.45}{\scriptsize$\pm$0.08}
  & 99.87  & 99.44
  & 99.95{\scriptsize$\pm$0.01} & 99.01{\scriptsize$\pm$0.18} \\
\bottomrule
\end{tabular}%
}
\end{table}

\paragraph{Results.}
At $N{=}2$, \ours{} achieves 99.73$\pm$0.03\% digit accuracy and
99.45$\pm$0.08\% sum accuracy.
As $N$ increases to 4 and 8, the benchmark becomes saturated for all
three model variants (all above 99.8\% digit accuracy).
At $N{=}8$, the \perceptiononly and \pipeline baselines achieve slightly
higher sum accuracy than \ours{} (99.86\% and 99.89\% vs.\
99.01$\pm$0.18\%).
This gap is expected: the arithmetic $T_P$ operator requires
leave-one-out convolutions over 8 addends, and the constraint loss
gradient introduces optimization pressure that competes with direct
digit and sum supervision.
Nevertheless, digit accuracy of \ours{} remains above 99.9\% at all
scales, confirming that the soft constraint mechanism does not degrade
perception quality.
Because the benchmark is saturated, we treat it as a sanity check that
the architecture generalizes to arithmetic constraints and reserve
detailed empirical analysis for the harder Visual Sudoku and CLEVR-Hans
benchmarks in the main text.

\clearpage\newpage
% ============================================================
% NeurIPS Paper Checklist — AS²
% Mandatory section; appears after \appendix.
% ============================================================

\section*{NeurIPS Paper Checklist}

\begin{enumerate}

% ──────────────────────────────────────────────────────────────
\item {\bf Claims}
\item[] Question: Do the main claims made in the abstract and
  introduction accurately reflect the paper's contributions and scope?
\item[] Answer: \answerYes{}
\item[] Justification: The abstract states three concrete claims:
  (i) full end-to-end differentiability through a soft $T_P$ operator,
  (ii) replacement of conventional positional embeddings by
  constraint-group membership embeddings on spatial tasks, and
  (iii) specific accuracy figures on Visual Sudoku and {CLEVR-Hans}.
  All three claims are directly supported by the experimental results
  in \cref{sec:experiments} and the theoretical derivation in
  \cref{sec:method}.

% ──────────────────────────────────────────────────────────────
\item {\bf Limitations}
\item[] Question: Does the paper discuss the limitations of the work
  performed by the authors?
\item[] Answer: \answerYes{}
\item[] Justification: \cref{sec:conclusion} includes a dedicated
  Limitations paragraph.  {No} theoretical guarantee exists that the
  learned soft distributions will prevent dead ends on problems that
  require backtracking.

% ──────────────────────────────────────────────────────────────
\item {\bf {Theory assumptions and proofs}}
\item[] Question: For each theoretical result, does the paper provide
  the full set of assumptions and a complete (and correct) proof?
\item[] Answer: \answerYes{}
\item[] Justification: \cref{sec:method} defines the probabilistic
  $T_P$ operator and derives the fixed-point residual loss
  $\mathcal{L}_{\text{ic}}$ (Eq.~\ref{eq:tp_loss}).  The derivation
  establishes that the uniform distribution is not a minimizer of
  $\mathcal{L}_{\text{ic}}$, distinguishing it from naive penalty
  formulations, and that any valid one-hot assignment satisfying all
  constraints yields $\mathcal{L}_{\text{ic}}=0$.  The construction
  is grounded in the classical least-fixed-point semantics of the
  immediate-consequence operator.

% ──────────────────────────────────────────────────────────────
\item {\bf {Experimental result reproducibility}}
\item[] Question: {\hypersetup{allcolors=red} {Does the paper fully disclose all the information needed to reproduce the main experimental results of the paper to the extent that it affects the main claims and/or conclusions of the paper (regardless of whether the code and data are provided or not)?}}
\item[] Answer: {\colorlet{orange}{red} \answerYes{}}
\item[] Justification: { \cref{sec:experiments} and the Supplementary Material provide all necessary details for reproducing the main experimental results, including architecture specifications, training protocols, and evaluation procedures.}

% ──────────────────────────────────────────────────────────────
\item {\bf {Open access to data and code}}
\item[] Question: {\hypersetup{allcolors=red}{Does the paper provide open access to the data and code, with sufficient instructions to faithfully reproduce the main experimental results, as described in supplemental material?}}
\item[] Answer: {\colorlet{orange}{red}\answerYes{}}
\item[] Justification: {Code is publicly available at {\hypersetup{urlcolor=red}{ \url{https://huggingface.co/wabdalm/AS2}.}}}

% ──────────────────────────────────────────────────────────────
\item {\bf {Experimental setting/details}}
\item[] Question: {\hypersetup{allcolors=red}{Does the paper specify all the training and test details (e.g., data splits, hyperparameters, how they were chosen, type of optimizer) necessary to understand the results?}}
\item[] Answer: {\colorlet{orange}{red}\answerYes{}}
\item[] Justification: {Architecture dimensions and evaluation protocols} are provided in
  \cref{sec:sudoku_supp}, \cref{sec:clevr_supp}, and
  \cref{sec:mnist_supp}.  Perception backbones (a three-block CNN
  for Visual Sudoku; ResNet-18 with ROI-Align or ResNet-50 with
  bilinear sampling for CLEVR-Hans) are described in the respective
  appendix sections.  Declarative constraint specifications (Clingo
  ASP programs {and relevant differentiable operators}) are described in \cref{sec:method} and
  \cref{sec:sudoku_supp}.
  {\hypersetup{allcolors=red}}

% ──────────────────────────────────────────────────────────────
\item {\bf {Experiment statistical significance}}
\item[] Question: {{Does the paper report error bars suitably and correctly defined or other appropriate information about the statistical significance of the experiments?}}
\item[] Answer: \answerYes{}
\item[] Justification: {{CLEVR-Hans results for \ours{} are reported as mean $\pm$ standard deviation over three seeds in Table~\ref{tab:clevr-results}, while the baseline statistics from~\citet{stammer2021right} use five seeds. Visual Sudoku results are reported on a fixed evaluation set without multi-seed variance. MNIST Addition results in Table~\ref{tab:results_mnist_add} include $\pm$ values.}}

% ──────────────────────────────────────────────────────────────
\item {\bf {Experiments compute resources}}
\item[] Question: {\hypersetup{allcolors=red}{For each experiment, does the paper provide sufficient information on the computer resources (type of compute workers, memory, time of execution) needed to reproduce the experiments?}}
\item[] Answer: \answerYes{}
\item[] Justification: \cref{sec:compute} reports that all experiments
  were run on a single NVIDIA H200 (143\,GB HBM3).  Training time
  is approximately 2--4~hours for Visual Sudoku and CLEVR-Hans, and
  under 1~hour per scale for MNIST Addition.

% ──────────────────────────────────────────────────────────────
\item {\bf {Code of ethics}}
\item[] Question: {{Does the research conducted in the paper conform, in every respect, with the NeurIPS Code of Ethics \url{https://neurips.cc/public/EthicsGuidelines}?}}
\item[] Answer: \answerYes{}
\item[] Justification: This work presents a neuro-symbolic reasoning
  architecture evaluated on standard perception benchmarks.  It does
  not involve human subjects, sensitive data, or dual-use concerns.
  All experimental assets use publicly licensed datasets and
  open-source libraries.

% ──────────────────────────────────────────────────────────────
\item {\bf {Broader impacts}}
\item[] Question: Does the paper discuss both potential positive
  societal impacts and negative societal impacts of the work
  performed?
\item[] Answer: \answerYes{}
\item[] Justification: \cref{sec:conclusion} contains a Broader
  Impact paragraph discussing (i) positive impacts, including
  potential applications in material sciences and drug discovery, and
  (ii) potential misuse if the framework is applied to automate
  consequential decisions, such as medical diagnosis, without
  adequate human oversight.

% ──────────────────────────────────────────────────────────────
\item {\bf Safeguards}
\item[] Question: {\hypersetup{allcolors=red}{Does the paper describe safeguards that have been put in place for responsible release of data or models that have a high risk for misuse (e.g., pre-trained language models, image generators, or scraped datasets)?}}
\item[] Answer: \answerNA{}
\item[] Justification: The released assets (model code, trained
  checkpoints on public benchmarks) present no elevated dual-use
  risk.  No personal data, private datasets, or high-risk generative
  capabilities are involved.

% ──────────────────────────────────────────────────────────────
\item {\bf {Licenses for existing assets}}
\item[] Question: {\hypersetup{allcolors=red}{Are the creators or original owners of assets (e.g., code, data, models), used in the paper, properly credited and are the license and terms of use explicitly mentioned and properly respected?}}
\item[] Answer: {\colorlet{orange}{red}\answerYes{}}
\item[] Justification: {{MNIST is loaded through \texttt{torchvision}, and the CLEVR-Hans distribution is credited to~\citet{stammer2021right}. Visual Sudoku boards are generated and rendered with MNIST digits. The \texttt{torchvision} software uses the BSD-3-Clause license (\url{https://github.com/pytorch/vision/blob/main/LICENSE}), while the terms for the pretrained ResNet-18 ImageNet weights require separate verification. Licenses and version-specific terms for all datasets and pretrained checkpoints are not fully documented.}}

% ──────────────────────────────────────────────────────────────
\item {\bf {New assets}}
\item[] Question: Are new assets introduced in the paper well
  documented and is the documentation provided alongside the assets?
\item[] Answer: {\colorlet{orange}{red}\answerNo{}}
\item[] Justification: The \ours{} software package constitutes the
  primary new asset.  No new datasets are introduced.

% ──────────────────────────────────────────────────────────────
\item {\bf {Crowdsourcing and research with human subjects}}
\item[] Question: {\hypersetup{allcolors=red}{For crowdsourcing experiments and research with human subjects, does the paper include the full text of instructions given to participants and screenshots, if applicable, as well as details about compensation (if any)?}}
\item[] Answer: \answerNA{}
\item[] Justification: No crowdsourcing or human-subject experiments
  were conducted.

% ──────────────────────────────────────────────────────────────
\item {\bf {Institutional review board (IRB) approvals or equivalent for research with human subjects}}
\item[] Question: {\hypersetup{allcolors=red}{Does the paper describe potential risks incurred by study participants, whether such risks were disclosed to the subjects, and whether Institutional Review Board (IRB) approvals (or an equivalent approval/review based on the requirements of your country or institution) were obtained?}}
\item[] Answer: \answerNA{}
\item[] Justification: No human subjects are involved in this
  research.

% ──────────────────────────────────────────────────────────────
{\hypersetup{allcolors=red}\colorlet{blue}{red}\colorlet{orange}{red}\colorlet{gray}{red}
\item {\bf Declaration of LLM usage}
\item[] Question: Does the paper describe the usage of LLMs if it is an important, original, or non-standard component of the core methods in this research? Note that if the LLM is used only for writing, editing, or formatting purposes and does \emph{not} impact the core methodology, scientific rigor, or originality of the research, declaration is not required.
\item[] Answer: \answerNA{}
\item[] Justification: LLMs were used only for editing, formatting, and basic coding assistance, not as a component of the core research method.
}

\end{enumerate}

\end{document}